\documentclass[10pt]{article}
\usepackage{amsmath, setspace}
\usepackage{graphicx}
\usepackage{natbib}
\usepackage{multicol}
\usepackage{lineno}
 \makeatletter       
 \renewcommand{\@biblabel}[1]{#1.}
 \makeatother
\makeatletter

\newenvironment{figurehere}
  {\def\@captype{figure}}
  {}
\makeatother

\begin{document}
%\doublespacing
\noindent
{\huge\textbf{Rabbit, toad, and the Moon: Can machine categorize them into one class?}}
\\
\\
\\
{\Large Daigo Shoji$^{1}$}\\
\\
%\fontsize{12pt}{24pt}\selectfont
1. Institute of Space and Astronautical Science, Japan Aerospace Exploration Agency (JAXA), Chuo-ku, Sagamihara, Kanagawa,
Japan. (shoji.daigo@gmail.com)\\
%$*$ Corresponding author: shoji.daigo@gmail.com
\\
\\
%\newpage
%\linenumbers*[1]
\begin{multicols}{2}
\section*{Abstract}
Recent machine learning algorithms such as neural networks can classify objects and actions in video frames with high accuracy. Here, I discuss a classification of objects based on basal dynamic patterns referencing one tradition, the link between rabbit, toad, and the Moon, which can be seen in several cultures. In order for them to be classified into one class, a basic pattern of behavior (cyclic appearance and disappearance) works as a feature point. A static character such as the shape and time scale of the behavior are not essential for this classification. In cognitive semantics, image schemas are introduced to describe basal patterns of events. If learning of these image schemas is attained, a machine may be able to categorize rabbit, toad, and the Moon as the same class. For learning, video frames that show boundary boxes or segmentation may be helpful. Although this discussion is preliminary and many tasks remain to be solved, the classification based on basal behaviors can be an important topic for cognitive processes and computer science.

\section{Introduction}
It has been widely acknowledged that neural networks can classify objects in images as humans do. For example, in ImageNet Large Scale Visual Recognition Challenge (ILSVRC), images have been classified into 1000 classes with more than 95\% accuracy. Furthermore, as well as static images, dynamic actions in video frames have also been classified using machine learning (e.g., Karapathy et al., 2014; Wu et al., 2015, 2016; Hou et al., 1017; Gu et al., 2018). Action recognition can be used to label video content (video classification), which is an important topic for the interaction between people and computers. In the video classification, subject (e.g., human, dog) and type of action (e.g., eat, walk) are classified, and short descriptions such as "dog runs" are given. Using object detection techniques, providing boundary boxes or segmentation of detected objects has also been performed (Hou et al., 1017; Gu et al., 2018). 

We (people) also classify both subject and action by seeing/detecting the target. For example, when we see a dog moving fast, this event is recognized as "a dog is running." In this case, the type of subject (i.e., dog) is classified independent of its action (i.e., run). However, humans can categorize objects from basal actions themselves either. Especially in the fundamental level of our culture, cognition, and imagination, this type of categorization has often been performed. One example can be seen in the link between rabbit, toad, and the Moon. Although their shape and species are primarily different, in several cultures, the Moon has been linked to rabbit and toad (Thuillard and Le Quellec, 2017; Thuillard, 2021).
As shown below, the main reason for this link is that they all have a basal behavior/action that they appear and disappear cyclically. 
Usually, this pattern is categorized as the different types of action. In the case of the Moon, the cyclic change of its shape is called waxing and waning, while the annual disappearance of the toad is hibernation. The frequent appearance of new rabbits is meant as birth. However, based on the basal patterns of behavior/action themselves (i.e., recurrence of appearance and disappearance), the three objects are categorized into one class under the label of "fertility” or “immortality.” Gilles Deleuze, a French philosopher, says, "there are greater differences between a plow horse or draft horse and a racehorse than between an ox and a plow horse" (Deleuze, 1988). Plow horse can be classified with ox rather than race horse due to agricultural behavior. To understand human cognition and artificial intelligence, we have to study this type of categorization. 

Here, I show a preliminary discussion on the relationship between the three objects and on the possibility for a machine to classify them into one category. 

\section{Mechanism of the link between rabbit, toad, and the Moon}
In several countries, it has been believed that there is a rabbit/hare on the Moon (Thuillard and Le Quellec, 2017; Thuillard, 2021). For example, in China, the rabbit on the Moon is called "jade rabbit," and the jade rabbit is an essential motif of the Moon (Fig. \ref{fig1}). This tradition is very old in Asian countries. We can see sentences on the rabbit on the Moon in the Indian manuscripts, Jaiminiya brahmana and Satapata brahmana, which are thought to be edited in $\sim$600 B.C (Eggeling, 1900; Bodewitz, 1973). 
\\
\\
~~

\begin{figurehere}
\centering
\includegraphics[width=6cm]{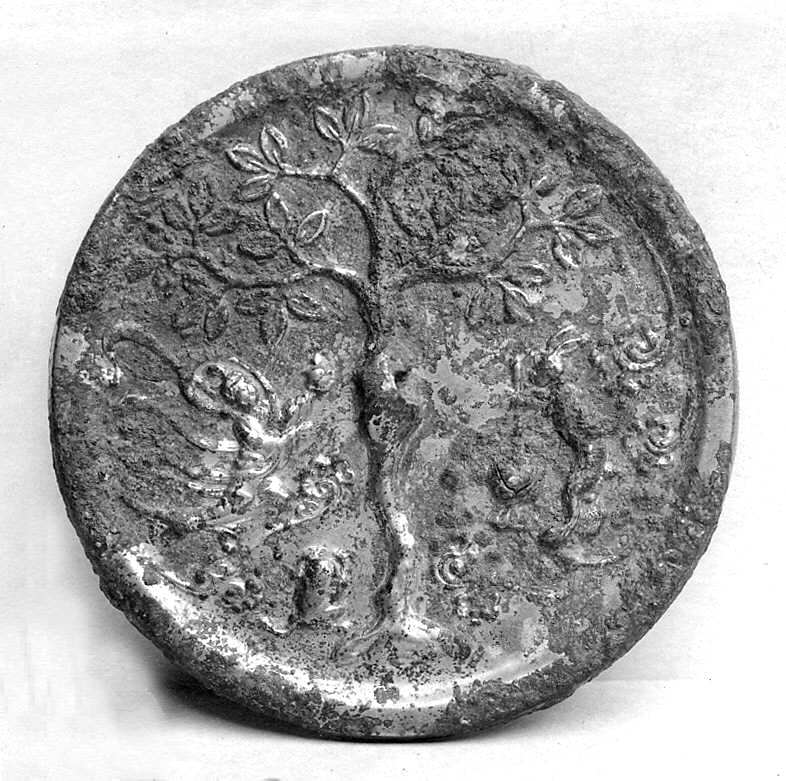}
\caption{Chinese bronze mirror which represents the Moon (7-8 century. The Metropolitan Museum of Art. Public domain image). On the right side of a tree, a rabbit pounds the elixir of immortality. A toad can be seen on the left side.}
\label{fig1}
\end{figurehere}
~~
\\

\begin{figure*}[htbp]
\centering
\includegraphics[width=12cm]{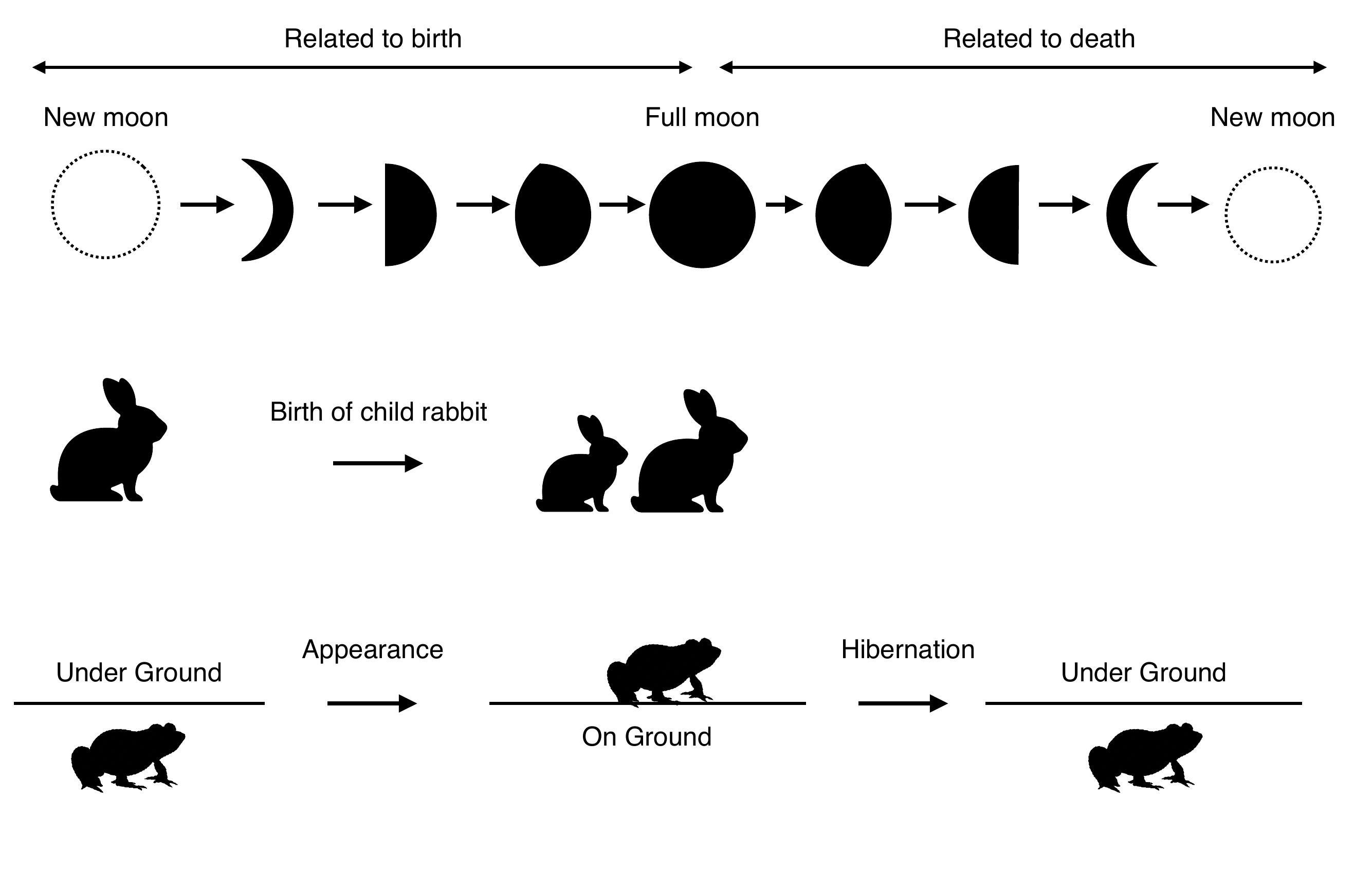}
\caption{Relationship between the Moon, rabbit, and toad. They all have a behavior of cyclic appearance and disappearance, which are related to birth and death, respectively. Although each action's type and time scale are different, the basal cyclic pattern categorizes them into one class. The categorized class is labeled as immortality or fertility.}
\label{fig2}
\end{figure*}

Typically, we explain that the reason for this culture is that the color pattern of the lunar surface is similar to the shape of a rabbit. Actually, the lunar surface has dark and bright areas caused by basaltic and anorthositic rocks. However, cultural anthropologists have indicated that the main reason the Moon is related to rabbit is that both are symbols of fertility or immortality (e.g.,  Thuillard and Le Quellec, 2017; Thuillard, 2021). We can see the cyclic change of the lunar shape, i.e., waxing and waning (Fig. \ref{fig2}). This appearance and disappearance of the Moon let ancient people imagine birth and death. Thus, based on this imagination, people began to think that the Moon experienced its birth and death many times. This recurrent behavior is also applied to rabbit because rabbit bears children frequently. Thus, both the Moon and rabbit were categorized into the same class that is labeled as "fertility" or "immortality" (recurrence of birth). Actually, in China, it has been believed that the Jade rabbit pounds elixir for immortality on the Moon (Fig. \ref{fig1}). In this link, the Moon and rabbit are classified into one category not by their shape (static state) but by their common behaviors (cyclic appearance). 

Along with rabbit, the Moon is also linked to toad in several countries (Thuillard, 2021). This connection can also be understood from their cyclic action of them because toad disappears by hibernation in winter and appears on the ground again in spring (Fig. \ref{fig2}). 

Although anthropologists discuss more complicated structures of this culture (e.g., Thuillard and Le Quellec, 2017; Thuillard, 2021), the critical point here is that, if we consider only static shape/state, rabbit, toad, and the Moon cannot be classified into the same category. Based on their dynamic behavior/action, they are connected so closely (Fig. \ref{fig2}). In the previous work, Shoji (2017) applied a convolutional neural network to know the similarity between the color pattern of the lunar surface and the shapes of several animals. However, only from the figures the relationship between rabbit, toad, and the Moon cannot be understood sufficiently. In this relationship, in contrast to attaching independently classified categories (e.g., after the categories of "dog" and "run" are recognized, then the event "a dog runs" is given), cyclic appearance and disappearance themselves makes the category that summarizes rabbit, toad, and the Moon. In other words, events are recognized like "one of the objects which run (move fast) is dog" rather than "a dog runs." In the former cognition, dogs may be linked to cars and even shooting stars if we are impressed with their movement. Although the word "run" may not be used for a shooting star, which is regarded as the usage of metaphor, the fundamental pattern of the action (i.e., moving fast) is common. 

The other important point for this kind of classification is that every element (e.g., rabbit, toad, the Moon) in one category does not always have a common character (Fig. \ref{fig2}). Rabbit is linked to the Moon by the cyclic appearance (i.e., the birth of a new rabbit and the Moon's waxing). On the other hand, in the case of the link between toad and the Moon, the cyclic disappearance as well as appearance is taken into account. The focused aspect of the action is a little bit different. Actually, the meanings of fertility and immortality are not the same. However, it is common that their appearance is related to birth. Thus, they can make one category based on birth and death. We can see a Chinese tradition that both rabbit and toad have lived together on the Moon since B.C. (Fig. \ref{fig1}). 
In the book “Patterns in comparative religion,” Mircea Eliade, a Romanian historian, shows that the Moon also makes relationships with many objects such as water, snake, and plant (Eliade, 1958). Plant growth has a cyclic pattern every year, which can be related to the change of the lunar shape (Fig. \ref{fig1}). In addition, Eliade says that water induces the germination of plants. Thus, water can also work as the cause of the cyclic appearance of the plant. Snake hibernates every year and molts, which gives rise to recurrence of birth and death. These relationships among elements are often not so clear. However, they make one categorical class as symbols of fertility and immortality (the recurrent birth and death). Ludwig Wittgenstein, an Austrian philosopher, called this kind of categorization “family resemblance” (Lakoff, 1987). In a family, every member does not always have one common character. While some people have a similar nose, others are similar in point of eyes or walking style. However, they are categorized as one family. 

\section{Application to machine learning}
As mentioned above, recent machine learning algorithms have succeeded in recognizing video actions (e.g., Karapathy et al., 2014; Wu et al., 2015, 2016; Hou et al., 1017; Gu et al., 2018). The question in this section is whether a machine can classify rabbit, toad, and the Moon into one category. 

The three objects are categorized as symbols of immortality or fertility. The meanings of immortality and fertility are based on birth and death. Both birth and death are the most impressive irreversible events for people. Thus, the recurrent appearance and disappearance of the Moon and toad became a symbol of immortality. In the case of the rabbit, ancient people must have been surprised that rabbit bears many children so often. Our lives and fundamental feelings, such as surprise, can be an origin for what kind of label is given. Thus, providing a label to the classified category seems difficult for a machine. 

However, how about the classification itself? As a neural network can classify objects from characteristic shapes, can a machine classify objects from distinct patterns of behavior? 
To answer this question, the following points are essential.\\
\\
1. The shape of each object is ignored. The only basal pattern of behavior is focused.\\

Every object has multiple behaviors. For example, in addition to waxing and waning, the Moon changes its position in the sky. Rabbit and toad move and jump. These motions are not used to make the relationship shown above. Only appearance, disappearance, and recurrence are critical actions in the link between rabbit, toad, and the Moon. Labels we usually call (e.g., waning and hibernation) are also not principal information. 
\\
\\
2. Actual timescale of each behavior is not an important factor. \\

For example, the period of lunar waxing and waning occurs in $\sim$30 days. On the other hand, the time scale of hibernation is one year. The interval of birth of rabbits is uncertain compared with the lunar cycle. When the critical action is focused, the actual time scale of each action is not considered. Thus, the length of time of each behavior should be normalized in classification. 
\\
\\
3. Objets are sometimes linked as family resemblance.\\

Each element in the same category does not always have a common character. Thus, we sometimes have to merge each category if two elements in different categories have common behavior. For example, water can be linked to the Moon because rain causes germination (i.e., the appearance of the plant).
\\

To consider these tasks, image schema plays an important role. Image schema describes the schematic pattern of action and state which people experience in their lives (Johnson, 1987; Lakoff, 1987;  Mandler and C\'anovas, 2014). In cognitive semantics, image schemas are used to analyze human cognition and language structure. For example, the usage of “in” and “out” can be understood from the schema of container (e.g., Johnson, 1987). Although these two words are used in various situations, we always have the structure of the schema when they are used. In metaphor, structures of image schemas are transported from one concept to another (e.g., Johnson, 1987; Lakoff, 1987). Due to this mechanism, we can make our obscure experiences clear. Johnson (1987) introduced several basic image schemas in his book “The body in mind,” and the pattern of cycle is one of them (Fig. \ref{fig3}). The image schema of cycle is represented as a rounded shape with arrows that indicate the direction of action/change. 

One important point of image schemas is that they do not show concrete actions. Image schemas describe abstracted event patterns in our world (Johnson, 1987). In other words, due to this extraction, many types of concrete recurrent actions/changes can be related to the schema of cycle. Johnson (1987) says that image schemas are generated by experiencing similar patterns and extracting the fundamental patterns from these experiences. The relationship between rabbit, toad, and the Moon can be understood by the schema of cyclic pattern. For example, ancient people recognized the recurrent appearance and disappearance of toads and the Moon, and these two objects are classified into one class as symbols of immortality. 
\\
\\
~~

\begin{figurehere}
\centering
\includegraphics[width=7cm]{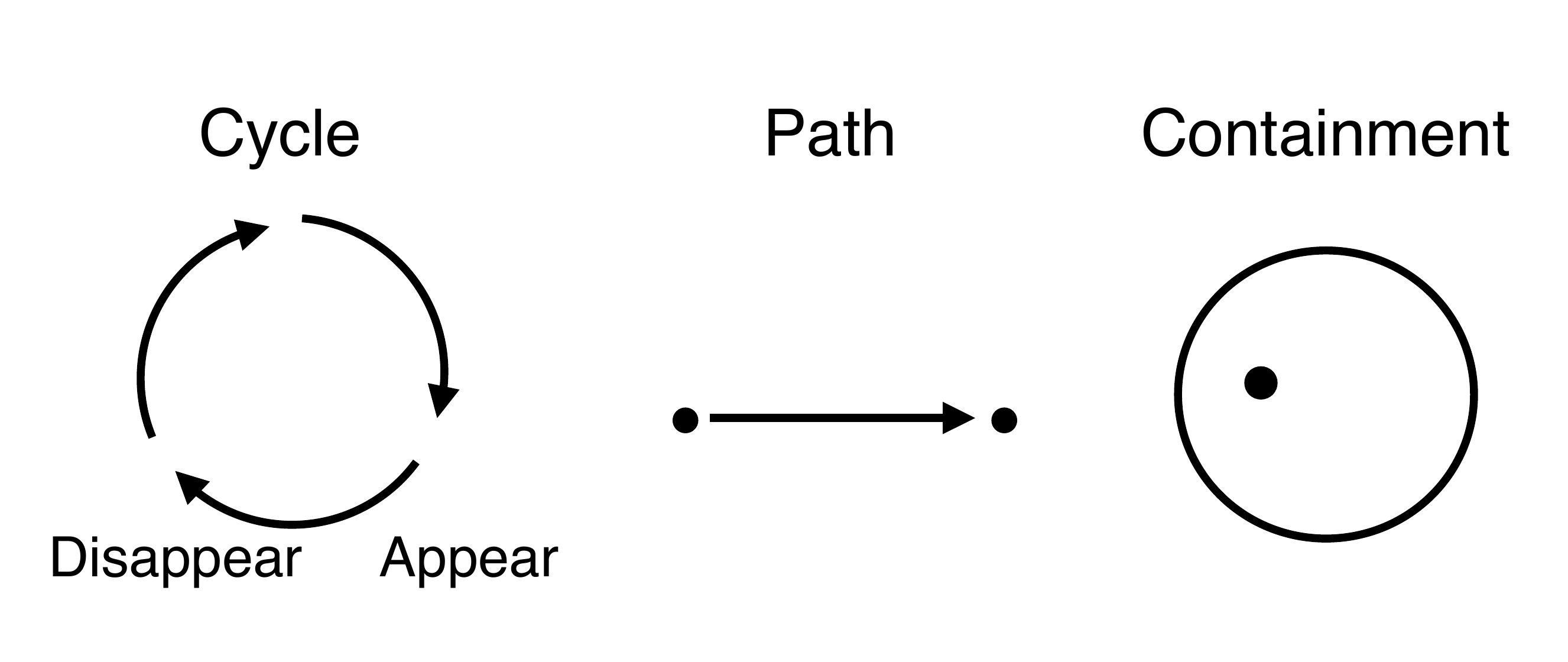}
\caption{Three examples (cycle, path, and container) of basal image schema suggested by Johnson (1987). In the schema of cycle, an action that a path can represent occurs recurrently. For the link between rabbit, toad, and the Moon, appearance and disappearance consist of cyclic patterns.}
\label{fig3}
\end{figurehere}
~~

However, for them to be the symbols of immortality, their appearance and disappearance must be regarded as birth and death in advance because their behavior is not actual birth and death. This structure can also be understood from the image schema. A recurrent event is achieved when the same action/change continues many times (Fig. \ref{fig3}). In the image schema of cycle, several arrows consist of a cycle (Fig. \ref{fig3}). Each arrow can be regarded as the image schema of path (Fig. \ref{fig3}). In the case of the Moon and toad, if we focus on one phase of cycle, they appear from nothing and disappear again (Fig. \ref{fig2}). Thus, it is not surprising that ancient people imagined a path from birth to death relating to their lives because it is the most important event. The relationship between rabbit and the Moon is also linked because rabbit bears children frequently, which is the exact birth. In this case, appearance is the schema of path. Although the focused event as the path is slightly different, as a family resemblance, they are classified as one category based on the recurrence of a similar path, which made rabbit and toad the animals on the Moon in several cultures.

\begin{figure*}
\centering
\includegraphics[width=17cm]{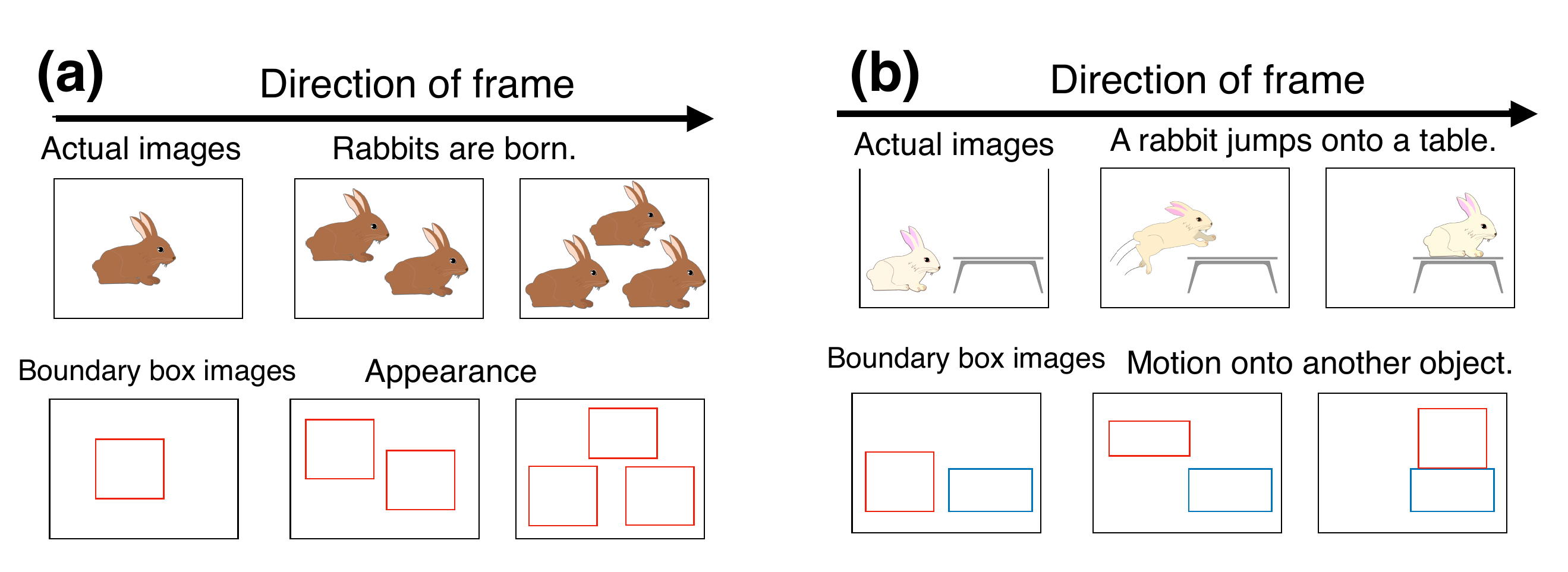}
\caption{Schematic view of learning image schemas by machine. (a): Schema of appearance. (b): Schema of onto. Current algorithms can detect and classify objects and concrete actions by learning video frames. Drawing boundary boxes or segmentation of detected objects has also been performed. After detecting many concrete actions, several basal patterns may be extracted by learning boundary boxes or segmentation in frames. Using the extracted actions and states, the detected objects can have abstracted characters  (e.g., rabbit--recurrent appearance, rabbit--the motion of onto). The common abstracted characters connect objects and organize into a new class (e.g., rabbit--recurrent appearance--the Moon). In this case, the label itself of the extracted action, such as "appearance" and "onto" is not important. Patterns shown in boundary boxes include key information. What kind of new label is provided (e.g., fertility) seems difficult for machines because labeling is based on people's lives and feelings.}
\label{fig4}
\end{figure*}

Is it possible for a machine to learn basal image schemas? As mentioned above, action recognition by machine learning has succeeded in detecting and classifying videos' actions. Thus, if we use video frames that show modeled motions or changes, learning schemas may be possible. Of course, many tasks must be discussed and solved for future studies. For example, the effects of shape and timescale must be removed. In the case of the image schema of cycle, the position where the pattern occurs is not an important factor. Thus, to extract only key patterns, we should use images that show several kinds of shapes and positions.

An alternative way may be using the boundary box itself to describe the relationship and action between objects. In object detection, the detected objects' area is usually shown with rectangle boxes. Because the shape of the bounding box is fixed to a rectangle, the configuration and action of bounding boxes in video frames may be able to represent several image schemas. For example, if a bounding box appears at a specific video frame, it shows the schema of appearance (Fig. \ref{fig4}). Considering several frames, recurrent appearance may also be learned. Of course, several schemas are difficult to describe by boundary box. For example, the configuration of "in" is difficult to discriminate from that of "in front of" because both cases are described as small boxes contained in a large box. Thus, applying segmentation method or stereo images can also be important. However, boundary box and the segmentation of video frames may have potential as the learning of image schemas. After a machine learns concrete actions, by training the frames that show only boundary box or segmentation, several image schemas may be extracted, which creates a relationship between the detected objects and the extracted schemas (Fig. \ref{fig4}). In other words, the boundary box or segmentation can show superordinate concepts (e.g., appearance, into, onto) of concrete actions. Due to this character, image schemas can work as classes, which makes new links between objects and re-organizes them. In these new links, a label of extracted schema is not important information. Action and configuration of boundary box or segmentation (i.e., the pattern itself) should be used to link each object. The Moon may be classified together with rabbit and toad in this way. However, further discussions and studies are required to construct a detailed algorithm.
 
To normalize the effect of timescale is also another task. Animals have many behaviors whose time scale is largely different. Observing for one year, hibernation can be seen while hopping finishes within a few seconds. Thus, we need to consider the proper time scale to extract basal behavior. If we focus on jumping, long-time observation is not appropriate. Like climate data (e.g., the amount of precipitation per day, per month, and per year), observational data with multiple scales should be considered. These days, time-lapse videos can be taken easily, which may help the learning of behaviors. 

\section{Summary}
In this work, I discussed how rabbit and toad are linked to the Moon in several countries and the possibility for a machine to learn this relationship. The most crucial point is that the Moon appears and disappears (i.e., waxing and waning). According to cultural anthropologists, this behavior of the Moon was related to birth and death. Toad also disappears in winter by hibernation and appears again in spring. This similarity of the fundamental behavior linked toad to the Moon as symbols of immortality. Rabbit is related to the Moon because it bears many children frequently, which is associated with the cyclic waxing of the Moon. This relationship cannot be understood well from their static figures. The basal pattern of their dynamic behaviors generated the links between them, and the three objects were categorized as symbols of immortality or fertility. For machines to classify them as one category, learning image schemas, which represent the basal dynamic action, may be helpful. Because the construction of image schemas is based on our lives and imagination, machines cannot learn every schema. Even for several image schemas, many tasks are remained to be solved. 

The classification based on action patterns is a fundamental cognition mechanism of people. In the early stage, the infant is more attracted to event rather than object itself (Mandler and C\'anovas, 2014). Thus, the discussion between image schemas and machine learning can be a very important topic in the field of cognitive semantics and AI. The relationship between rabbit, toad, and the Moon is not only the topic of folklore and anthropology but also the topic of current people.

%\section*{Acknowledgements}
%\addcontentsline{toc}{section}{Reference}
\nocite{*}

%\bibliographystyle{plain}
%\bibliography{article}
%\bibliographystyle{newapa}

\end{multicols}
\end{document}